\documentclass[conference]{IEEEtran}
\IEEEoverridecommandlockouts
\usepackage{cite}
\usepackage{amsmath,amssymb,amsfonts}
\usepackage{algorithmic}
\usepackage{graphicx}
\usepackage{textcomp}
\usepackage{xcolor}
\def\BibTeX{{\rm B\kern-.05em{\sc i\kern-.025em b}\kern-.08em
    T\kern-.1667em\lower.7ex\hbox{E}\kern-.125emX}}
\begin{document}

\title{Turkish Text Classification: From Lexicon Analysis to Bidirectional Transformer\\
{\footnotesize \textsuperscript{}}
}

\author{\IEEEauthorblockN{Deniz Kavi}
\IEEEauthorblockA{ \textit{The Koc School}\\
Istanbul, Turkey \\
denizk2022@stu.kocschool.k12.tr}
\and
\IEEEauthorblockN{\textsuperscript{}}
\IEEEauthorblockA{\textit{} \\\\
 \\
}
}

\maketitle

\begin{abstract}
Text classification has seen an increased use in both academic and industry settings. Though rule based methods have been fairly successful, supervised machine learning has been shown to be most successful for most languages, where most research was done on English. In this article, the success of lexicon analysis, support vector machines, and extreme gradient boosting for the task of text classification(and sentiment analysis) are evaluated in Turkish and a pretrained transformer based classifier is proposed, outperforming previous methods for Turkish text classification. In the context of text classification, all machine learning models proposed in the article are domain-independent and don’t require any task-specific modifications.
\end{abstract}

\section{Introduction}

\subsection{Task Description}
The fundamental problem of text classification is determining the label of a sample text such as a tweet or a movie review. Text classification can be approached both as a supervised learning problem and an unsupervised learning problem. In unsupervised learning the model would be given a dataset of texts without any label and the model would be tasked with grouping texts based on similarities etc. And for a supervised learning problem the model would be given texts with labels and the model would optimize its predictions based on the given labels. In our case, supervised methods will be used as the benchmark dataset is fully labeled. Sentiment analysis can be treated as a sub-task of text classification as sentiments can be represented as different classes from texts.

\subsection{Datasets}
The selected benchmark dataset was the Turkish Movie Sentiment dataset used in Demirtas, E. and Pechenizkiy, M. Cross-lingual Polarity Detection with Machine Translation\cite{b1}. The dataset features 5331 examples for both positive and negative sentiments of movie Turkish reviews and the model will be tasked to determine whether the review is positive and negative. 

\subsection{Models Overview}
This article will be going over the supervised learning models of Support Vector Machines, Extreme Gradient Boosting\cite{b2} and a BERT\cite{b3}\cite{b4} transformer based classifier. It will also include a review of lexicon and supervised learning methods from Gezici et al. \cite{b5}

\section{Related Studies}
Non-deep learning machine learning approaches to text classification can directly be applied to Turkish as they do not treat words or language through "understanding". In other words, gradient boosting or support vector machines can directly be applied to data regardless of the text's language.\\

In Gezici et. al\cite{b5}, the authors propose combining a combination of supervised learning methods along with a polarity lexicon. Meaning that a polarity lexicon is created and the polarity of a set of words is defined before the model is trained on the dataset. This approach reaches an accuracy of 78 percent on the benchmark sentiment analysis(and text classification) dataset. \\

Since computers are unable to "understand" words, the words in the inputted sentences are to be vectorized through word embeddings. Word embeddings are created in a way that the embeddings of words are closer to each other in a 2D coordinate plane. Where mathematical operation can be done on words' embeddings: \\

\begin{equation}
\begin{aligned}
Parents = Family - Children \\
Ebeveyn = Aile - Cocuklar
\end{aligned}
\end{equation}
FastText employs word vectors, which are obtained both through supervised and unsupervised methods\cite{b6}\cite{b7}. The library's developers provide models for 294 languages and word embeddings for 157 languages. The primary concern besides having a high accuracy, is to have a short training time. \\

Although convolutional neural networks were initially designed to be used for images; convolutions, max pooling and other ways of approaching image classification problems are applicable to text classification tasks. As regular CNN architectures take 2D images as input, they are modified to fit for 1 dimensional text data.  \\

Character-level Convolutional Networks for Text Classification\cite{b8} proposes a CNN model that trains on characters instead of words or sentences. Instead of word embeddings etc. they encode every character in the target language's alphabet, but this approach lacks in crucial data for language understanding. \\

ULMFiT\cite{b9} was one of the first methods to propose transfer learning for language tasks. It pioneered the idea of training a language model on a corpus of English and fine-tuning on the sentence level. The proposed method is inspired by ULMFIT, though moving from a AWD LSTM language model to a BERT language model.  \\

\section{Methods}

\subsection{Support Vector Machine (SVM)}

\paragraph*{Model Description}

SVMs are supervised machine learning models mostly used for classification and regression. Since sentiment analysis is a classification problem this description will focus on SVMs for classification. For p dimensional data(2 in this case) The SVM training algorithm constructs a (p-1) dimensional hyperplane. The SVM attempts to maximize the separation, or margin between the 2 classes. This is known as a linear classifier. \\

\paragraph*{Application}

The hyperparameters for the SVC were:
\begin{itemize}
  \item Regularization parameter: 1
  \item Cache size: 200
  \item Decision function: one vs. rest

  \item Kernel Coefficient: “scale”
  \item Radial Basis Function Kernel 
\end{itemize}

A count vectorizer was used to convert text to a matrix of token counts. A TF-IDF Vectorizer was used to get a matrix of TF-IDF features. \\

\paragraph*{Results}

The SVM application was the worst performing out of the 3 supervised machine learning approaches in this article. It could only produce an accuracy of approximately 50 percent. However, because the base model underperformed compared to most other models, hyperparameter optimization was not attempted, which would likely improve the results.

\subsection{Extreme Gradient Boosting(XGBoost)}

\paragraph*{Model Description}
XGBoost has been one of the most successful models for tabular/structured data problems. Tens of academic competitions have been won by teams using XGBoost, though unfortunately it is unable to reach the success of deep learning based models for text classification. 

\paragraph*{Boosting} 
Simple tree based models are the building block for boosting systems. These tree based predictors are intentionally only slightly better than randomly guessing, this is why they are known as weak learners. In a boosting model such as XGBoost, these weak learners are combined to form a singular strong learner, which makes the final prediction. \\

For weak learner h(x) giving prediction [-1, +1]:  
Strong learner H(x) would make a prediction:

\begin{equation}
H(x) = h_{1}(x)+h_{2}(x)+h_{y}(x) 
\end{equation}

\paragraph*{Application}\label{AA}
The hyperparamaters for the model are as follows:

\begin{itemize}
  \item Subsampling by Tree: 0.9
  \item Column Sampling by Tree: 0.7
  \item Minimum Loss Reduction: 0.1
  \item Maximum depth of each tree: 6  
  \item Minimum Child Weight:1 
  \item Learning Rate: 0.3  

\end{itemize}

A count vectorizer and a TF-IDF vectorizer were also used in the XGBoost model for feature extraction. Stop words(words that such as the or a that are assumed not have any meaning) were removed.

\paragraph*{Results}

The XGBoost model had a 93 percent training accuracy and a validation accuracy of 88 percent. Although it doesn't perform as well as its deep learning and transformer-based counterparts, it is much faster to train. The XGBoost model trains only on the benchmark dataset without needing any pretraining on a large corpus of the target language and compared to a BERT model, it trains much faster on the benchmark dataset.

\subsection{BERT Transformer Classifier}
\paragraph*{Model Description}

Bidirectional Encoder Representation from Transformers(BERT) is a language representation model. Using pretrained language models(which is BERT in this case) and fine-tuning them on classification datasets have been shown by a large pool of works\cite{b9}\cite{b4} as more successful than their previous counterparts. There a tangible improvements in model performance when using BERT and other pretraining-based, transfer learning architectures.

In this article, the BERT architecture is modified to fit a text classification task, further explained in the application section. \\
\subsubsection*{Recurrent Neural Networks(RNNs)}
Before transformers, RNNs and Long-Short Term Memory based models were the preferred neural network architecture for most natural language processing tasks. RNNs are neural network architectures that model statistical relationships using sequence data. The task of text classification can be represented as a sequence to vector problem, where the model take a sequence(text) as input and return a vector of probabilites for the labeled classes. A sequence to sequence model could be explained with a machine translation task, where the model would recieve a sequence in one language and return a sequence in a different language. \\

\hspace{10mm}Although RNNs are fairly successful in these tasks they come with unique drawbacks. The problem of vanishing and exploding gradients arise when the model is working with long sequences. A solution to this problem was proposed with the Long-Short Term Memory architecture. \\

\subsubsection*{Long-Short Term Memory(LSTM)}  
LSTMs\cite{b6} are a type of RNN architecture which introduced a cell called a Long Short Term Memory cell which allowed for models to retain longer sequences in memory. Although better than simple RNNs, LSTM cells were still unable to reach required performance.  \\

\subsubsection*{Transformer}
A transformer neural network\cite{b7} is an encoder-decoder system similar to RNNs, though instead of needing to process words individually, in a transformer architecture, the model takes in words in parallel and generates word embeddings in parallel. \\
\subsubsection*{Input Embeddings}
As computers are unable to understand raw text or words, word inputs must be converted to vectors or matrices. Word embeddings are exist in an embedding space, where coordinates of words that are related in meaning are closer together. \\

\begin{figure}[htp]
    \centering
    \includegraphics[width=10cm]{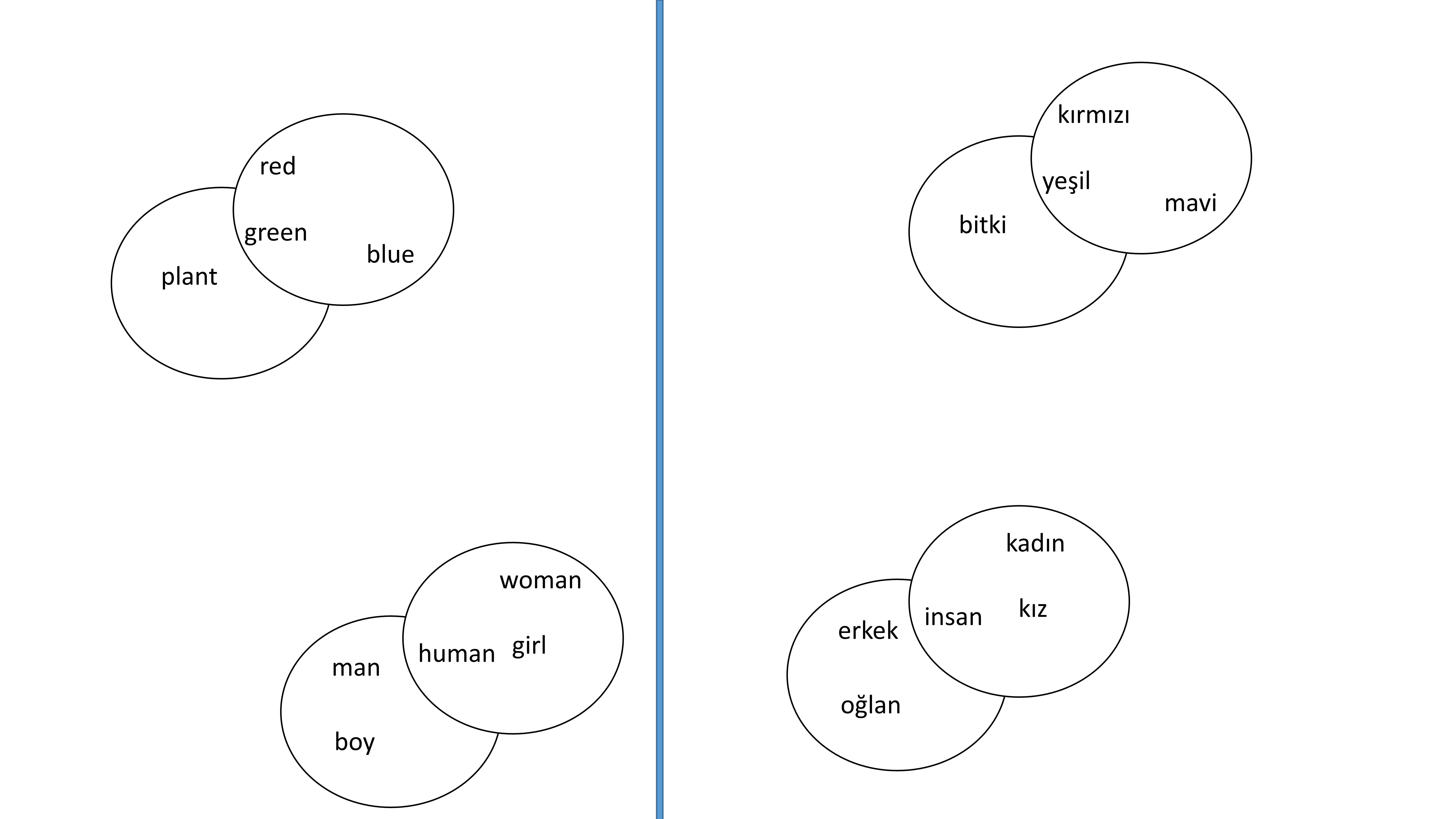}
    \caption{Turkish and English Embedding Example}
    \label{fig:galaxy}
\end{figure}

\subsubsection*{Positional Encoders}
As words' meanings change based on their position in sentences we will have to design the model to understand what the word means in context. A positional gives context based on where the word is in the sentence. \\
\subsection*{Transformer to BERT}
The components of a transformer described above are used for the encoding of words. BERT's architecture is composed of a stack of encoders, unlike the original transformer architecture, BERT doesn't only work in the machine translation domain. It is first, (pre)trained on a large corpus of the language and then fine-tuned for the specific task, which includes question answering, text classification, and text summarization. For a BERT model to achieve sufficient results in language understanding the task can be represented by Masked Language Modeling and Next Sentence prediction. For the former, the model would be predicting what words to fill the blanks with such as: \\

The boy wasn't [MASK] enough to drive. or: \\ 
\hspace{10mm}Onun [MASK] araba kullanmak icin cok kucuk. \\ 

In this case, the model would be predicting a word from the sentence which has been replaced with the token [MASK]. For a Next Sentence Prediction task, the model would be given two examples of sentences and asked to determine if one comes after the other. BERT trains on these tasks simultaneously.\cite{b3} Fine-tuning for the text classification task is further explained in the application section. 

\subsection*{Application}
The Turkish BERT base model from the Huggingface transformers library \cite{b8}, which was itself provided as a "community model" by the Bavarian State Library. It contains 12 encoder(transformer) blocks and a hidden size of 768. Its maximum number of tokens per input is 512. A classifier with a softmax layer is added on top of the BERT model to predict the probability of a label. A pretraining approach on the sentence-level was proposed in ULMFiT\cite{b9} was also implemented for Turkish as part of the fast.ai MOOC. The pretrained BERT model obtained from the Huggingface Transformers community was initially trained on a large Turkish corpus and then retrained on the benchmark text classification/sentiment analysis dataset. The model will be made available to be used with the transformers library on Github. \footnote{https://github.com/denizkavi/turkish-bert-classifier} \\

\begin{figure}[htp]
    \centering
    \includegraphics[scale=0.025]{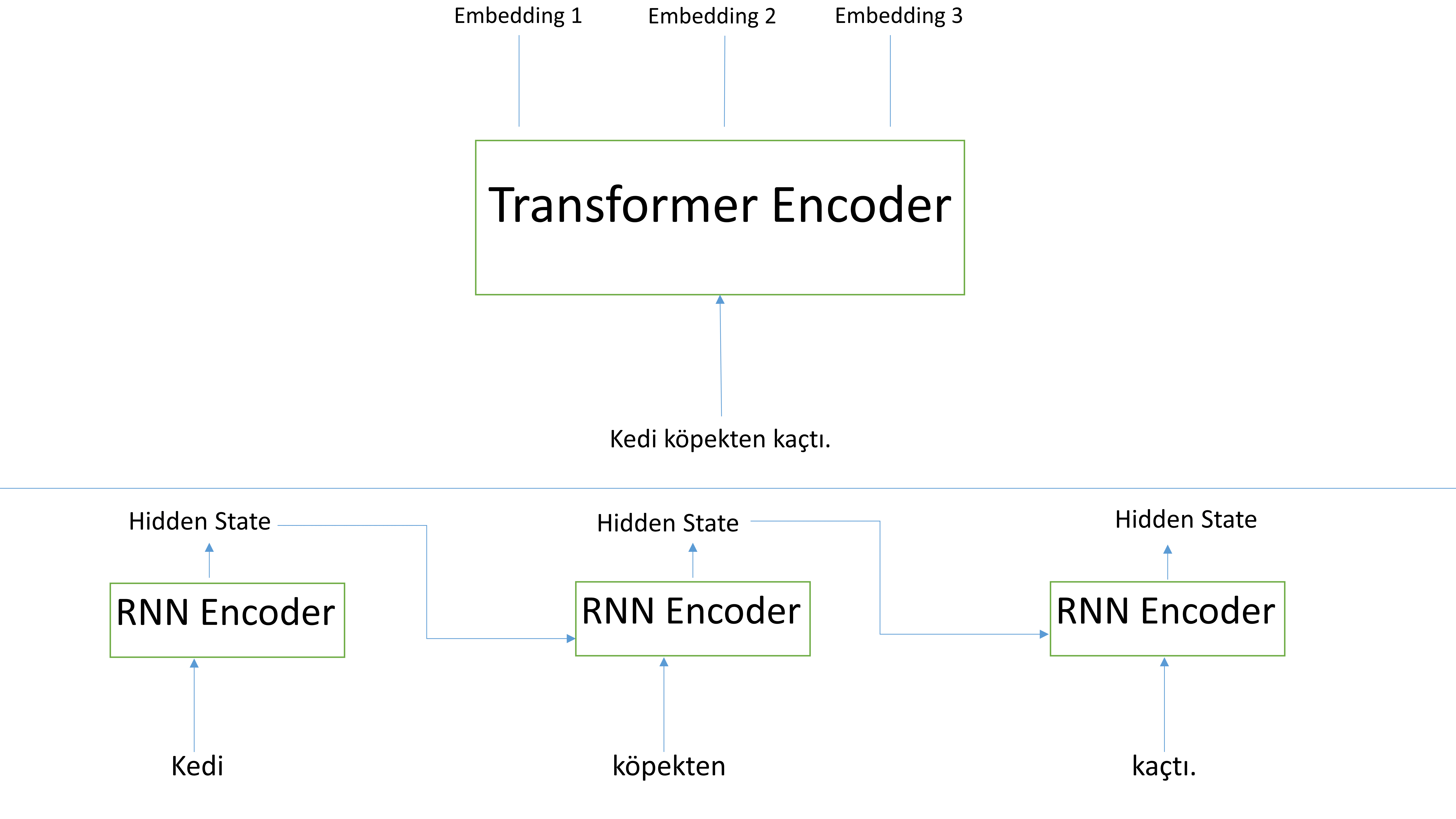}
    \caption{Differences in encoding between RNNs and transformers}
    \label{fig:encoder}
\end{figure}

In the ULMFiT\cite{b9} paper, the authors proposed novel methods to improve classifier model performance using pretraining. Namely discriminative fine-tuning, slanted triangular learning rates, and gradual unfreezing. \\

\subsubsection*{Discriminative Fine-Tuning} 
In discriminative fine-tuning, instead of implementing a  single learning rate for all layers of the model but to tune a different learning rate to every layer. The authors intuitively  found that it is best to choose a learning rate for the last layer, $n_{l}$, where the learning rate for the layer before would be calculated by:
$n_{l-1} = n_{l}/2.6 $. \\
\subsubsection*{Slanted Triangular Learning Rate(STLR)}
STLRs are learning rates that linearly increase and then linearly decay, in accordance to the update schedule. \\
\subsubsection*{Gradual Unfreezing}
As fine-tuning every layer of the model at once may cause catastrophic forgetting, ULMFiT proposes unfreezing layers from the last layer to the first layer. Every time a layer is unfrozen, it will be fine-tuned for one epoch. \\
\subsubsection*{Hyperparameters}
The pretrained Turkish BERT model\cite{b10} was trained on a Tesla K80 GPU with a batch size of 16, the maximum sequence length of 500, a starting learning rate of 4e-5, also using STLRs, gradual unfreezing and discriminative fine-tuning. Stop words were not removed. 

\subsection{Results}
The pretrained BERT Transformer model outperforms previous Turkish text classification and sentiment analysis methods with a validation accuracy of 92.5 percent and a training accuracy of 99 percent. However, it does take a long time to train compared to more traditional machine learning approaches, especially when a language model pretrained on a large corpus dataset isn't available. \\

\section*{Conclusion}

After an overview of existing approaches, this paper proposes a BERT architecture with an added softmax layer which achieves state-of-the-art performance on the dataset with an accuracy of 92.5 percent. Although fine-tuning a base BERT model on the dataset for a text classification/sentiment analysis task was found to be tangibly more successful than earlier attempts, training times compared to non-deep learning approaches are much longer for a BERT model. 

\begin{table}[htbp]
\caption{Model Performances}
\begin{tabular}{|c|c|}
\hline
Model Name & Validation Accuracy(Percentage) \\
\hline
Support Vector Machine & 50 \\
\hline
Polarity Lexicon\cite{b5} & 78 \\
\hline
Extreme Gradient Boosting & 88 \\
\hline
ULMFiT\cite{b11} & 91 \\
\hline
BERT & 92.5 \\
\hline
\end{tabular}
\end{table}

\end{document}